\documentclass[10pt,logo,copyright]{nvidiatechreport}
\usepackage[authoryear,sort&compress,round]{natbib}

\usepackage[utf8]{inputenc} 
\usepackage[T1]{fontenc}    
\usepackage{xcolor}
\usepackage{parskip}        
\usepackage{url}            
\usepackage{booktabs}       
\usepackage{amsfonts}       
\usepackage{nicefrac}       
\usepackage{microtype}      
\usepackage{xcolor}         
\usepackage[dvipsnames]{xcolor} 
\usepackage{graphicx}
\usepackage{animate}        
\usepackage{subcaption}
\usepackage{tabularx}
\usepackage{makecell}
\usepackage{adjustbox}
\usepackage{setspace}
\newcolumntype{M}[1]{>{\centering\arraybackslash}m{#1}}
\usepackage{float}
\usepackage{tikz}
\usetikzlibrary{positioning,shapes,arrows}
\usepackage{amsmath,amsfonts,bm, bbm,leftindex}
\usepackage{multirow}
\usepackage{comment}
\usepackage{gensymb}
\usepackage{lipsum}
\usetikzlibrary{arrows.meta, positioning, fit}
\usepackage[para]{threeparttable}
\usepackage{tikz}
\usetikzlibrary{tikzmark}
\usepackage{caption}
\def\eqref#1{equation~\ref{#1}}

\def\1{\bm{1}}

\DeclareMathAlphabet{\mathsfit}{\encodingdefault}{\sfdefault}{m}{sl}
\SetMathAlphabet{\mathsfit}{bold}{\encodingdefault}{\sfdefault}{bx}{n}

\makeatletter
\let\save@mathaccent\mathaccent
\newcommand*\if@single[3]{%
  \setbox0\hbox{${\mathaccent"0362{#1}}^H$}%
  \setbox2\hbox{${\mathaccent"0362{\kern0pt#1}}^H$}%
  \ifdim\ht0=\ht2 #3\else #2\fi
  }
\newcommand*\rel@kern[1]{\kern#1\dimexpr\macc@kerna}
\newcommand*\widebar[1]{\@ifnextchar^{{\wide@bar{#1}{0}}}{\wide@bar{#1}{1}}}
\newcommand*\wide@bar[2]{\if@single{#1}{\wide@bar@{#1}{#2}{1}}{\wide@bar@{#1}{#2}{2}}}
\newcommand*\wide@bar@[3]{%
  \begingroup
  \def\mathaccent##1##2{%
    \let\mathaccent\save@mathaccent
    \if#32 \let\macc@nucleus\first@char \fi
    \setbox\z@\hbox{$\macc@style{\macc@nucleus}_{}$}%
    \setbox\tw@\hbox{$\macc@style{\macc@nucleus}{}_{}$}%
    \dimen@\wd\tw@
    \advance\dimen@-\wd\z@
    \divide\dimen@ 3
    \@tempdima\wd\tw@
    \advance\@tempdima-\scriptspace
    \divide\@tempdima 10
    \advance\dimen@-\@tempdima
    \ifdim\dimen@>\z@ \dimen@0pt\fi
    \rel@kern{0.6}\kern-\dimen@
    \if#31
      \overline{\rel@kern{-0.6}\kern\dimen@\macc@nucleus\rel@kern{0.4}\kern\dimen@}%
      \advance\dimen@0.4\dimexpr\macc@kerna
      \let\final@kern#2%
      \ifdim\dimen@<\z@ \let\final@kern1\fi
      \if\final@kern1 \kern-\dimen@\fi
    \else
      \overline{\rel@kern{-0.6}\kern\dimen@#1}%
    \fi
  }%
  \macc@depth\@ne
  \let\math@bgroup\@empty \let\math@egroup\macc@set@skewchar
  \mathsurround\z@ \frozen@everymath{\mathgroup\macc@group\relax}%
  \macc@set@skewchar\relax
  \let\mathaccentV\macc@nested@a
  \if#31
    \macc@nested@a\relax111{#1}%
  \else
    \def\gobble@till@marker##1\endmarker{}%
    \futurelet\first@char\gobble@till@marker#1\endmarker
    \ifcat\noexpand\first@char A\else
      \def\first@char{}%
    \fi
    \macc@nested@a\relax111{\first@char}%
  \fi
  \endgroup
}
\makeatother

\definecolor{darkred}{rgb}{0.7, 0.0, 0.0}

\usepackage{pifont}

\usepackage[nameinlink]{cleveref}
\crefname{equation}{Eq.}{Eqs.}
\crefname{figure}{Fig.}{Figs.}
\crefname{section}{Sec.}{Sec.}
\crefname{appendix}{App.}{App.}
\crefname{table}{Tab.}{Tabs.}
\crefname{algorithm}{Algo}{Algo}
\crefname{thm}{Thm}{Thm}
\Crefname{thm}{Thm}{Thm}
\crefname{prop}{Prop}{Prop}

\newcommand{\crefnames}[3]{%
  \@for\next:=#1\do{%
    \expandafter\crefname\expandafter{\next}{#2}{#3}%
  }%
}

\title{Industrial cuVSLAM Benchmark \& Integration }

\author[1]{Charbel Abi Hana}
\author[1]{Kameel Amareen}
\author[1]{Mohamad Mostafa}
\author[2]{Dmitry Slepichev}
\author[2]{Hesam Rabeti}
\author[2]{Zheng Wang}
\author[2]{Mihir Acharya}
\author[1]{Anthony Rizk}

\affil[1]{Idealworks}
\affil[2]{NVIDIA}

\begin{abstract}
This work presents a comprehensive benchmark evaluation of visual odometry (VO) and visual SLAM (VSLAM) systems for mobile robot navigation in real-world logistical environments. We compare multiple visual odometry approaches across controlled trajectories covering translational, rotational, and mixed motion patterns, as well as a large-scale production facility dataset spanning approximately 1.7 km. Performance is evaluated using Absolute Pose Error (APE) against ground truth from a Vicon motion capture system and a LiDAR-based SLAM reference. Our results show that a hybrid stack combining the cuVSLAM front-end with a custom SLAM back-end achieves the strongest mapping accuracy, motivating a deeper integration of cuVSLAM as the core VO component in our robotics stack. We further validate this integration by deploying and testing the cuVSLAM-based VO stack on an NVIDIA Jetson platform.
\end{abstract}

\begin{document}
\maketitle
\abscontent

\section{Introduction}
Autonomous navigation in warehouse and logistics environments presents persistent challenges for robotic systems. Long repetitive aisles, dynamic obstacles (moving forklifts, pallets, and personnel), and the absence of rich structure expose fundamental limitations of LiDAR-based particle filter localization; the dominant paradigm in mobile robotics; leading to localization drift and map inconsistencies. 

Visual Simultaneous Localization and Mapping (VSLAM) offer a compelling alternative. Camera-based systems capture rich photometric and structural information, enabling more robust feature extraction where 1D LiDAR measurements fall short. Critically, visual modalities inherently encode semantic cues; shelf labels, floor markings, signage; that can anchor localization in otherwise repetitive spaces, making VSLAM a well-suited paradigm for the structured yet visually diverse nature of logistics facilities. 

Despite this promise, practical VSLAM performance in real industrial deployments remains insufficiently characterized. Existing benchmarks focus primarily on controlled indoor environments or outdoor driving scenarios, leaving a gap in understanding system behavior under real warehouse industrial conditions: variable lighting, frequent occlusions, semi-dynamic scene changes, and long-duration operation. 

Beyond accuracy, industrial deployment imposes strict computational constraints. AMRs are not dedicated perception platforms; they concurrently run motion planning, obstacle avoidance, hardware coordination and intercommunication processes, all competing for CPU resources on embedded hardware. A VSLAM front-end that saturates the CPU risks starving safety-critical processes and breaking real-time guarantees. This demands a localization solution that is both highly accurate and computationally lightweight by design. 

GPU-accelerated, modular VSLAM architectures are uniquely positioned to meet both demands simultaneously. By offloading the full perception pipeline; feature extraction, stereo matching, pose estimation and optimization; to dedicated GPU hardware, the CPU is freed for higher-level autonomy tasks. NVIDIA's cuVSLAM embodies this philosophy: a CUDA-accelerated, modular VSLAM library delivering high-frequency, low-latency odometry and mapping on NVIDIA Jetson platforms, the embedded compute backbone of modern robotic fleets. Its architectural separation of front-end and back-end further enables flexible integration into heterogeneous robotics stacks without sacrificing accuracy or determinism. 

This paper addresses that gap with a systematic benchmarking study of leading VSLAM approaches; RTAB-Map \cite{labbe2019rtab}, cuVSLAM \cite{korovko2025cuvslam}, ORB-SLAM3 \cite{campos2021orb}, and an in-house hybrid method combining visual odometry from NVIDIA cuVSLAM with Idealworks’ SLAM backend; evaluated on data collected in real logistics environments. We assess both visual odometry and full SLAM performance, providing quantitative and qualitative comparisons across accuracy and robustness, with the goal of offering practitioners actionable insight into the readiness of current VSLAM methods for industrial deployment. 
\section{Benchmarked Systems and Evaluation Criteria}
\label{sec:methods}

The AMR is equipped with a LIPSEdge AE400 IR Stereo Camera, which provides synchronized stereo image pairs at a resolution of 1280x720 pixels and a frame rate of 15 FPS. The camera uses a rolling shutter technology, which can introduce motion blur during rapid movements, further challenging visual odometry and SLAM performance.

\subsection{Benchmarked Systems}

\paragraph{ORB-SLAM3}
ORB-SLAM3~\cite{campos2021orb} is a feature-based SLAM system supporting monocular, 
stereo, and RGB-D configurations. Its key contribution is the SLAM Atlas, a multi-map 
representation enabling robust relocalization and seamless merging of disconnected 
mapping sessions, complemented by a high-recall place recognition algorithm that 
enforces geometric consistency.

\paragraph{RTAB-Map}
RTAB-Map~\cite{labbe2019rtab} is a sensor-agnostic graph-based SLAM library supporting 
visual, 2D/3D LiDAR, and hybrid configurations. Its distinguishing feature is a memory 
management mechanism that bounds loop closure computation time, enabling scalable 
long-term operation in large environments.

\paragraph{cuVSLAM}
cuVSLAM~\cite{korovko2025cuvslam} is a CUDA-accelerated SLAM system designed for 
real-time pose estimation on edge devices. It supports arbitrary monocular, RGBD \& stereo multi-camera 
configurations; from a single camera up to 32 synchronized units; with a fully 
GPU-accelerated pipeline from feature extraction to bundle adjustment.

\subsection{Evaluation Metric}

Performance is quantified using Absolute Pose Error (APE), which measures the 
pose-wise translational deviation between an estimated trajectory 
$\mathbf{p}_e = \{p_e^i\}_{i=1}^{n}$ and its ground truth 
$\mathbf{p}_r = \{p_r^i\}_{i=1}^{n}$:
\begin{equation}
    \text{APE} = \frac{1}{n}\sum_{i=1}^{n} \|p_e^i - p_r^i\|_2
\end{equation}
Prior to evaluation, estimated trajectories are aligned to the ground truth via 
the Kabsch-Umeyama similarity transformation~\cite{88573} to account for scale 
and orientation differences. All evaluations use the EVO toolkit~\cite{grupp2017evo}.
\section{Visual Odometry Benchmark}
\label{sec:vo_benchmark}

\subsection{Benchmark Dataset}

\begin{figure}[htbp]
  \centering
    \includegraphics[width=0.9\textwidth]{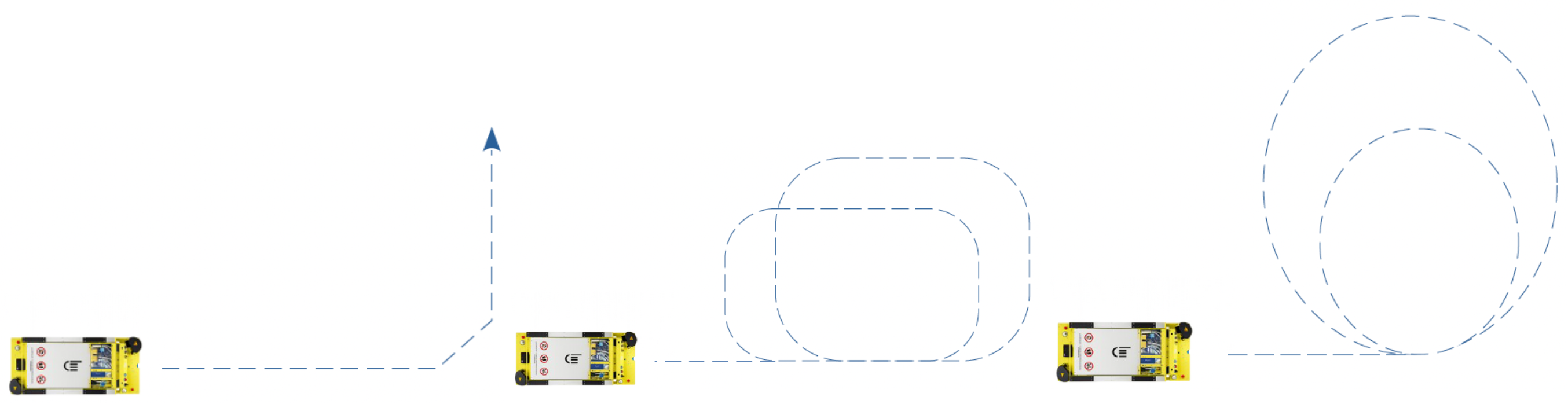}
    \caption{VO trajectories (from left to right): L-shape, hybrid \& rotation-heavy sequences.}
    \label{fig:vo_trajectories}
\end{figure}

The VO benchmark was conducted in an R\&D warehouse facility using three 
ROS2 bags, each capturing a distinct trajectory designed to isolate and 
stress different aspects of visual odometry performance, as illustrated 
in Figure~\ref{fig:vo_trajectories}.

The \textbf{L-shaped} sequence begins with a translational segment to calibrate 
the tracker, followed by an incremental rotation up to 90\textdegree, and 
concludes with a second translational segment; isolating rotation-induced drift 
and its compounding effect on subsequent translation. The \textbf{hybrid} sequence 
combines a long translational segment with mixed translation-rotation movements 
to evaluate drift accumulation over extended paths. The \textbf{rotation-heavy} 
sequence stress-tests the system under predominantly rotational motion with minimal 
translation. Ground truth for all sequences is obtained from a high-precision 
motion capture system. Trajectory statistics are summarized in 
Table~\ref{tab:vo_trajectory_analysis}.

\begin{table}[h]
\centering
\caption{Trajectory motion statistics per VO sequence.}
\label{tab:vo_trajectory_analysis}
\resizebox{\textwidth}{!}{%
\begin{tabular}{lcccc}
\toprule
\multirow{2}{*}{\textbf{Sequence}}
  & \multicolumn{2}{c}{\textbf{Linear Velocity (m/s)}} 
  & \multicolumn{2}{c}{\textbf{Angular}} \\
\cmidrule(lr){2-3} \cmidrule(lr){4-5}
  & \textbf{Mean} & \textbf{Max} 
  & \textbf{Mean Angular Vel. (deg/s)} & \textbf{Max Angular Acceleration (deg/s$^2$)} \\
\midrule
L-shaped       & 0.53 & 2.05 & 6.44  & 16.69 \\
Rotation-Heavy & 0.31 & 1.00 & 29.33 & 37.80 \\
Hybrid  & 0.42 & 2.76 & 25.35 & 39.28 \\
\bottomrule
\end{tabular}}
\end{table}

\subsection{Experimental Setup}

All three methods are evaluated in a stereo-only VO configuration. Wheel encoder 
odometry is included as a reference baseline, as it remains the dominant odometry 
source in logistics robotics despite its susceptibility to long-term drift. Each 
configuration is evaluated over 10 independent runs per sequence; ORB-SLAM3 is 
evaluated with a single run as its keypoint sampling produces deterministic output. Runs are reproduced on both an x86 system or \textbf{Desktop} which comprises an Intel i9 13th generation CPU with dual RTX Titan V GPUs and on an edge system which is the NVIDIA Jetson Xavier or \textbf{Xavier}.
Results are reported as mean APE across runs, with hyperparameters tuned to each 
method's best-performing configuration.

\subsection{Results and Analysis}

\begin{figure}[h!]
  \centering
    \includegraphics[width=0.7\textwidth]{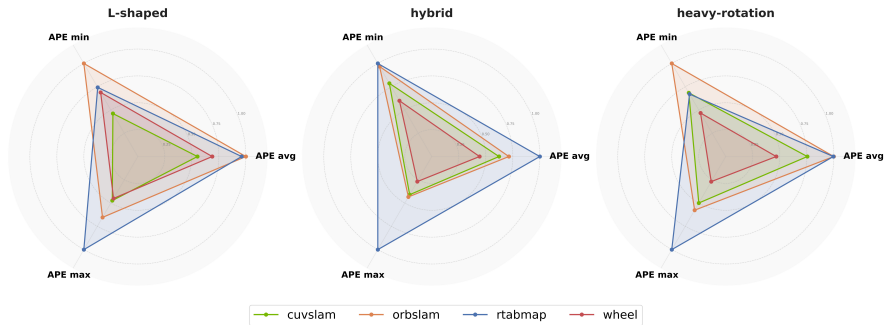}
    \caption{Radar plot comparing the metrics of each method across the three VO sequences. The wheel encoder baseline is included for reference. Smaller areas indicate better performance.}
    \label{fig:vo_results}
\end{figure}

Table~\ref{tab:vo_ape_results_extended} and Figure~\ref{fig:vo_results} summarize results across all three sequences on the \textbf{Desktop} setup. cuVSLAM consistently achieves the lowest APE among visual odometry methods across all scenarios. By offloading feature extraction and tracking to the GPU, cuVSLAM achieves an average inference time of 2.262 ms; roughly 10× faster than ORB-SLAM (21.828 ms) and 15× faster than RTAB-Map (35.033 ms); while maintaining the lowest average CPU. This GPU acceleration not only reduces per-frame latency but also frees CPU resources for other onboard processes, making cuVSLAM particularly well-suited for resource-constrained robotic platforms. 

The same conclusions hold for the \textbf{Xavier} setup, with marginally worse results attributable to its more constrained computational resources.

The wheel encoder baseline outperforms all visual methods on the rotation-heavy and hybrid sequences, primarily due to motion blur under rotational motion; a known limitation of our camera setup. 

\begin{table}[h]
\centering
\caption{APE (m) \& Computational Costs results across scenarios.
         APE: Mean $\pm$ std over 10 runs (single run for ORB-SLAM and Wheel Encoder) on \textbf{Desktop}}
\label{tab:vo_ape_results_extended}
\resizebox{\textwidth}{!}{%
\begin{tabular}{lcccccccc}
\toprule
\textbf{Method} & \textbf{L-shaped} & \textbf{Rotation-Heavy} & \textbf{Long Sequence} 
  & \textbf{Avg.\ Inference} & \textbf{Max Inference} 
  & \textbf{Avg.\ CPU} & \textbf{Max CPU} & \textbf{AVG GPU} \\
 & & & & \textbf{Time (ms)} & \textbf{Time (ms)} & \textbf{Usage (\%)} & \textbf{Usage (\%)} & \textbf{Usage (\%)} \\
\midrule
cuVSLAM       & $\mathbf{0.057 \pm 0.01}$  & $\mathbf{0.296 \pm 0.025}$ & $\mathbf{0.850 \pm 0.034}$
              & $\mathbf{2.262}$ & $\mathbf{3.137}$  & $\mathbf{2.002}$ & $8.600$ & $6.200$ \\
ORB-SLAM      & $0.103$       & $0.392$       & $0.974$
              & $21.828$ & $30.074$ & $4.759$ & $11.500$ & $-$\\
RTAB-Map      & $0.10 \pm 0.02$            & $0.390 \pm 0.14$           & $1.37 \pm 0.406$
              & $35.033$ & $45.354$ & $3.573$ & $\mathbf{7.200}$  & $-$\\
\midrule
Wheel Encoder & $0.071$       & $0.183$       & $0.604$
              & $-$ & $-$ & $-$ & $-$ & $-$\\
\bottomrule
\end{tabular}}
\end{table}

\begin{table}[h]
\centering
\caption{APE (m) \& Computational Costs results across scenarios.
         APE: Mean $\pm$ std over 10 runs (single run for ORB-SLAM and Wheel Encoder) on \textbf{Xavier}}
\label{tab:vo_ape_results_extended_xavier}
\resizebox{\textwidth}{!}{%
\begin{tabular}{lcccccccc}
\toprule
\textbf{Method} & \textbf{L-shaped} & \textbf{Rotation-Heavy} & \textbf{Long Sequence} 
  & \textbf{Avg.\ Inference} & \textbf{Max Inference} 
  & \textbf{Avg.\ CPU} & \textbf{Max CPU} & \textbf{AVG GPU} \\
 & & & & \textbf{Time (ms)} & \textbf{Time (ms)} & \textbf{Usage (\%)} & \textbf{Usage (\%)} & \textbf{Usage (\%)} \\
\midrule
cuVSLAM       & $\mathbf{0.067 \pm 0.05}$ & $\mathbf{0.301 \pm 0.12}$ & $\mathbf{0.640 \pm 0.43}$ & $\mathbf{9.400}$ & $\mathbf{40.340}$ & $\mathbf{9.601}$ & $61.011$ & $5.300$ \\
ORB-SLAM      & $0.150$ & $0.308 $ & $0.899$ & $60.342$ & $71.572$ & $36.009$ & $61.500$ & $-$\\
RTAB-Map      & $0.11 \pm 0.06$ & $0.311 \pm 0.17$ & $0.92 \pm 0.54$ & $86.300$ & $144.7$ & $25.600$ & $\mathbf{40.500}$ & $-$\\
\midrule
Wheel Encoder & $0.071$       & $0.183$       & $0.604$
              & $-$ & $-$ & $-$ & $-$ & $-$\\\bottomrule
\end{tabular}}
\end{table}

\section{Visual SLAM Benchmark}
\label{sec:vslam_benchmark}

\subsection{Dataset}

The VSLAM benchmark was conducted in the production facility of an external customer, where the robot navigated two predetermined paths: a primary sequence of approximately 1.7 km over 38 minutes of continuous operation, and a secondary sequence of approximately 570 m recorded in the same facility. Both paths contain more than 50 loop closure opportunities at roughly equal intervals. The mean linear speed of 0.64 m/s (max 2.01 m/s) and low mean angular activity of 1.93 deg/s reflect paths dominated by long straight segments with gradual turns; consistent with a warehouse aisle layout and ensuring that any localization failures are attributable to perceptual and map-consistency challenges rather than motion-induced degradation. All visual stacks share the same stereo input, feeding both VO front-ends and SLAM backends. 

Ground truth poses were obtained from the robot's onboard LiDAR-based localization system. To maximize ground truth accuracy, dynamic objects were cleared from the environment during the robot's path traversal, ensuring clean and reliable reference poses. For the VSLAM evaluation runs, the robot navigated the same predetermined path under nominal operating conditions. Runs exhibiting sporadic or abnormal behavior were discarded to maintain consistency across evaluations. This setup allows the results to reflect realistic deployment performance while remaining reproducible and comparable across stacks. 

\subsection{Experimental Setup}

Each configuration is evaluated over a single run along the predetermined paths. 
Four configurations are compared: \textbf{RTAB-Map} (RTAB-Map with
stereo VO and its own back-end); \textbf{cuVSLAM} (end-to-end stereo SLAM); 
\textbf{ORB-SLAM3} (end-to-end stereo SLAM); and the \textbf{Idealworks NVIDIA hybrid} (cuVSLAM 
stereo front-end with our SLAM back-end). Performance is assessed using APE 
against the LiDAR-based ground truth. These tests have been performed on the \textbf{Xavier} to emulate real world computational limitations under heavy SLAM stacks. We'll report the tick-rate of each approach measuring the number of localization poses outputted per second. In our AMR use-case, we can reasonably accept any system that is able to run at higher than $1Hz$, however different applications may have different constraints.

\subsection{Results and Analysis}

Table~\ref{tab:vslam_results} summarizes APE results for each method across 
both sequences. On Seq-570, RTAB-Map consistently lost visual odometry and 
failed to produce valid estimates, reflecting the fragility of its dense 
stereo front-end under the perceptual conditions of this facility. Across 
both sequences, cuVSLAM's stereo VO front-end proves to be the backbone of 
the strongest results. The Idealworks NVIDIA hybrid achieving the lowest 
mean APE of $0.91$~m on Seq-1700 with a tightly bounded RMSE of $0.54$~m. 
Maintaining sub-meter mean accuracy over a $1.7$~km trajectory demonstrates 
that cuVSLAM's front-end provides sufficiently consistent odometry for the 
back-end to close loops reliably and correct accumulated drift. The 
end-to-end cuVSLAM pipeline achieves the best mean APE of $0.32$~m on 
Seq-570, with higher drift on Seq-1700 at $1.57$~m expected in the absence 
of a dedicated map management back-end. By contrast, ORB-SLAM3 achieves 
$0.34$~m on Seq-570 and $1.17$~m on Seq-1700, and RTAB-Map yields $5.89$~m 
on Seq-1700, both substantially behind the cuVSLAM-powered configurations.

Regarding tick-rate, cuVSLAM's GPU acceleration is particularly evident: 
the end-to-end pipeline operates at $105.3$~Hz, well above the $1$~Hz 
operational threshold, while the Idealworks NVIDIA hybrid runs at $1.50$~Hz, 
sufficient for our use-case while dedicating most of its compute budget to 
back-end optimization. RTAB-Map falls to $0.95$~Hz, dropping below the 
$1$~Hz criterion and disqualifying it for our AMR use-case. The CPU overhead 
further underscores cuVSLAM's advantage: whereas ORB-SLAM3 consumes 
$52.28\%$ average CPU and peaks at $79\%$, the Idealworks NVIDIA hybrid 
requires only $18.74\%$ on average despite running a full SLAM back-end, a 
direct consequence of offloading the perception front-end entirely to the 
GPU and freeing CPU resources for safety-critical onboard processes.

These results position cuVSLAM as the critical enabling technology in this 
benchmark. As a standalone pipeline it delivers best-in-class throughput and 
strong accuracy, and as the front-end of the Idealworks NVIDIA hybrid it 
underpins the strongest overall localization performance. Its stability also 
stands in direct contrast to ORB-SLAM3, which exhibits sporadic segmentation 
faults that are unacceptable where continuous uptime is a hard requirement. 
It is also worth noting that cuVSLAM was bottlenecked by our rolling shutter 
sensor at 15~FPS; its ideal operating conditions involve a global shutter 
camera at 60 or even 100~FPS, so these results represent a conservative 
lower bound on its full capability.

\begin{figure}[h!]
    \centering
    \begin{minipage}[c]{0.54\textwidth}
        \centering
        \includegraphics[width=0.75\textwidth]{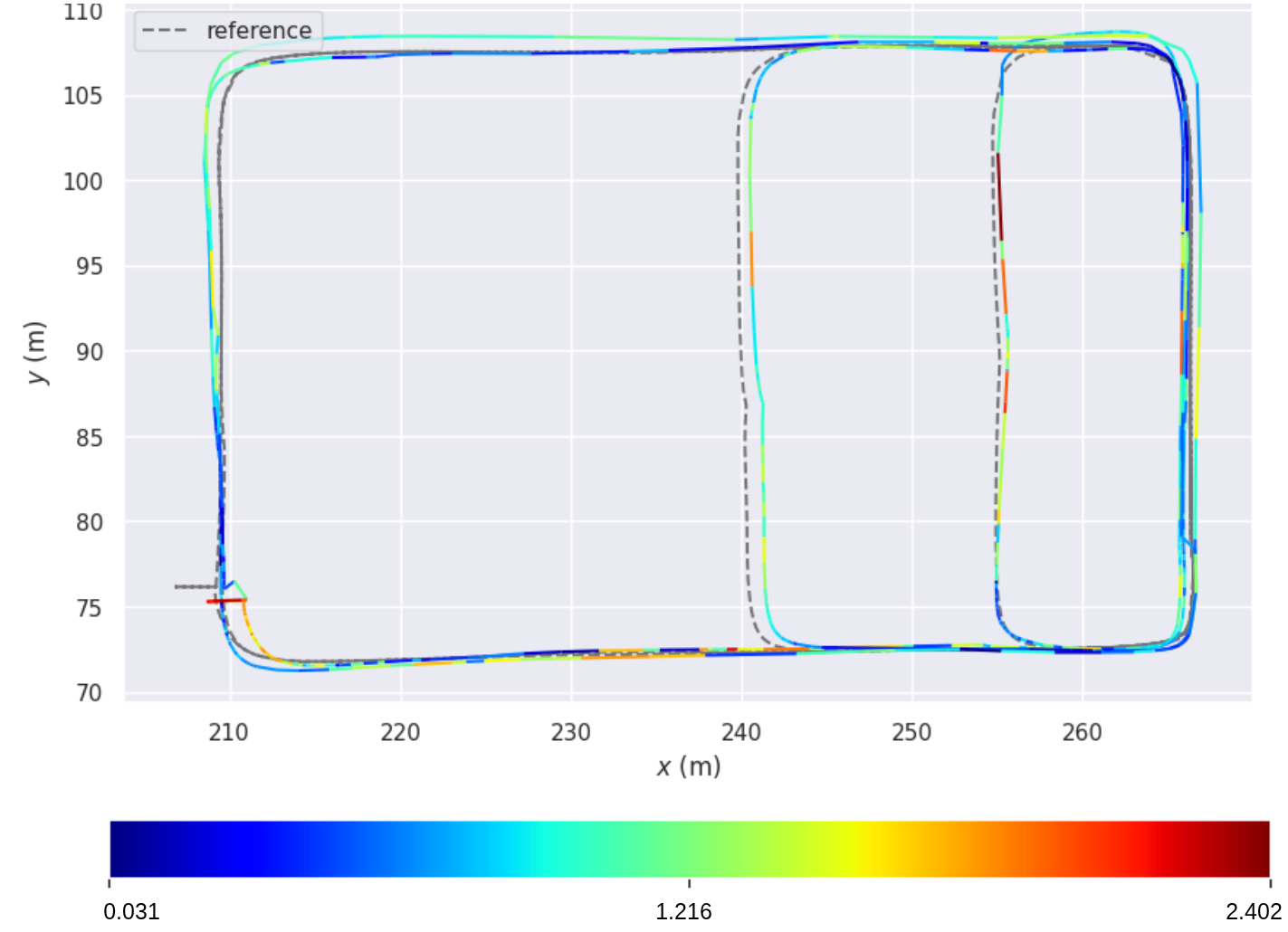}
        \subcaption{Estimated trajectory errors on Seq-570.}
        \label{fig:vslam_trajectories_570}
    \end{minipage}
    \hfill
    \begin{minipage}[c]{0.43\textwidth}
        \centering
        \includegraphics[width=\textwidth, 
            height=0.5\textheight,
            keepaspectratio]{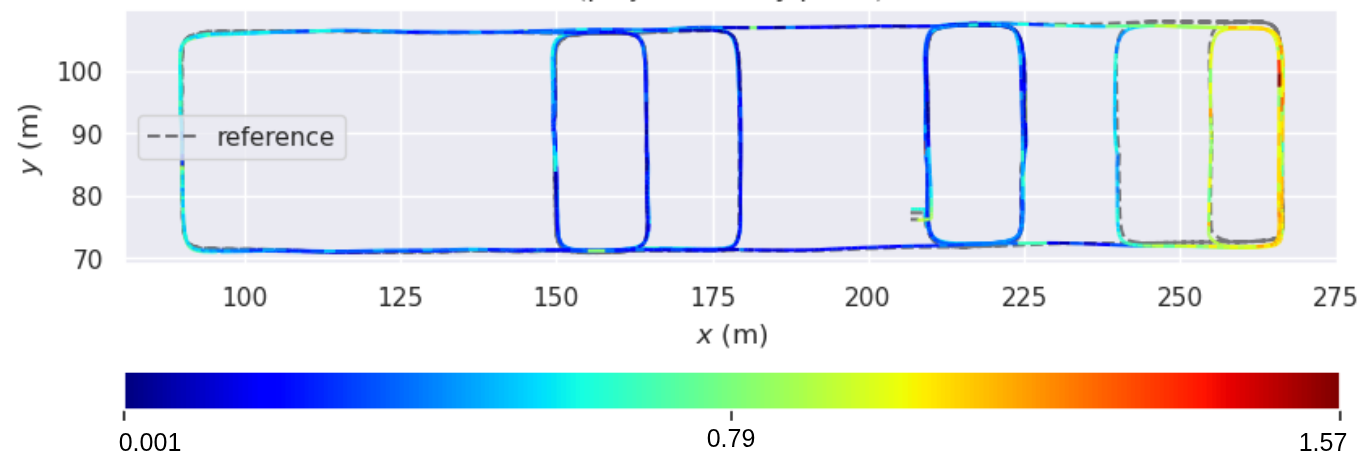}
        \subcaption{Estimated trajectory errors on Seq-1700.}
        \label{fig:vslam_trajectories_1700}
    \end{minipage}
    \caption{Idealworks NVIDIA hybrid estimated trajectory errors across both sequences.}
    \label{fig:vslam_trajectories}
\end{figure}

Figure~\ref{fig:vslam_trajectories} visually confirms the consistency and 
accuracy of the cuVSLAM-powered trajectory estimates across both sequences, 
with the ground truth trajectory shown as dashed lines for reference.

Table~\ref{tab:vslam_deployability} provides a qualitative deployability comparison 
across the four evaluated methods. The Idealworks NVIDIA Hybrid ranks first overall, 
combining the computational efficiency of the cuVSLAM GPU-accelerated front-end with 
post-run map management and multi-session as well as multi-robot mapping support; 
capabilities that are essential for long-term operation in shared logistics environments. 
cuVSLAM ranks second, offering broad sensor modality support and consistently low 
inference times, but lacks post-run map management tooling, which limits its 
out-of-the-box suitability for large-scale deployments. RTAB-Map ranks third: its 
memory management system enables stable operation on unbounded maps and natively 
supports multi-robot and multi-session scenarios, but its dense processing stack yields 
slower and at times unstable odometry, as corroborated by its failure on Seq-570. 
ORB-SLAM ranks last despite competitive accuracy; the absence of large-map handling, 
a pure localization mode, and multi-session support, combined with degrading inference 
times as the map grows and the significant development effort required to reach 
production readiness, make it the least viable candidate for real-world deployment.

\begin{table}[h]
\centering
\caption{VSLAM APE (m) results on the deployment dataset on \textbf{Xavier}}
\label{tab:vslam_results}
\resizebox{\textwidth}{!}{%
\begin{tabular}{lcccccc c c c c}
\toprule
\multirow{2}{*}{\textbf{Method}} 
  & \multicolumn{3}{c}{\textbf{Seq-570}} 
  & \multicolumn{3}{c}{\textbf{Seq-1700}} 
  & \multirow{2}{*}{\textbf{\shortstack{Tick-Rate \\ (Hz)}}}
  & \multirow{2}{*}{\textbf{\shortstack{Avg. CPU \\ Usage (\%)}}}
  & \multirow{2}{*}{\textbf{\shortstack{Max CPU \\ Usage (\%)}}}
  & \multirow{2}{*}{\textbf{\shortstack{Avg. GPU \\ Usage (\%)}}} \\
\cmidrule(lr){2-4} \cmidrule(lr){5-7}
  & \textbf{Mean} & \textbf{Max} & \textbf{RMSE} 
  & \textbf{Mean} & \textbf{Max} & \textbf{RMSE} & & & & \\
\midrule
cuVSLAM                  & $\mathbf{0.32 \pm 0.25}$ & $1.27$ & $0.41$ & $1.57 \pm 1.25$ & $5.21$  & $2.01$ & $\mathbf{105.3}$ & $\mathbf{9.10}$ & $\mathbf{33.00}$ & $\mathbf{65.00}$ \\
Idealworks NVIDIA Hybrid & $0.61 \pm 0.29$ & $1.09$ & $0.68$ & $\mathbf{0.91 \pm 0.29}$ & $\mathbf{2.2}$ & $\mathbf{0.54}$ & $1.50$ & $18.74$ & $35.00$ & $66.26$ \\
RTAB-Map                 & $-$             & $-$    & $-$    & $5.89 \pm 4.69$ & $22.14$ & $7.53$ & $0.95$ & $-$ & $-$ & $-$ \\
ORB-SLAM                 & $0.34 \pm 0.22$ & $1.61$ & $0.38$ & $1.17 \pm 1.16$ & $5.19$  & $1.65$ & $12.50$ & $52.28$ & $79.00$ & $-$ \\
\bottomrule
\end{tabular}}
\end{table}

\begin{table}[h]
\centering
\caption{Deployability comparison of VSLAM methods.}
\label{tab:vslam_deployability}
\resizebox{\textwidth}{!}{%
\begin{tabular}{lp{4.5cm}p{4.5cm}p{4cm}c}
\toprule
\textbf{Method} 
  & \textbf{Large-Scale Support} 
  & \textbf{Inference Time} 
  & \textbf{Production Grade} 
  & \textbf{Rank} \\
\midrule
Idealworks NVIDIA Hybrid
  & \textcolor{green}{$\oplus$} Post-run map management \newline
    \textcolor{green}{$\oplus$} Multi-session \& multi-robot mapping support
  & \textcolor{green}{$\oplus$} GPU-accelerated pipeline powered by cuVSLAM \newline
    \textcolor{green}{$\oplus$} Consistent inference time with growing map size
  & \textcolor{green}{$\oplus$} Production-ready \newline
    \textcolor{green}{$\oplus$} Seamless ROS2 integration
  & 1\textsuperscript{st} \\
\midrule
cuVSLAM
  & \textcolor{green}{$\oplus$} Multi-camera support \newline
    \textcolor{green}{$\oplus$} Multi-modality support (IMU, RGB-D, Stereo) \newline
    \textcolor{red}{$\ominus$} No post-run map management tools
  & \textcolor{green}{$\oplus$} GPU-accelerated \newline
    \textcolor{green}{$\oplus$} High throughput from SLAM stack
  & \textcolor{green}{$\oplus$} Production-ready \newline
    \textcolor{green}{$\oplus$} ROS2 integration via Isaac ROS Visual SLAM
  & 2\textsuperscript{nd} \\
\midrule
RTAB-Map
  & \textcolor{green}{$\oplus$} Memory management system \newline
    \textcolor{green}{$\oplus$} Constant inference time on unbounded maps \newline
    \textcolor{green}{$\oplus$} Multi-robot \& multi-session support
  & \textcolor{red}{$\ominus$} Dense stack, generally slower \newline
    \textcolor{red}{$\ominus$} Unstable odometry under rapid motion \newline
    \textcolor{orange}{$\sim$} Inference time stable with map size
  & \textcolor{green}{$\oplus$} Near production-ready \newline
    \textcolor{orange}{$\sim$} Moderate integration effort
  & 3\textsuperscript{rd} \\
\midrule
ORB-SLAM3
  & \textcolor{red}{$\ominus$} No large-map handling \newline
    \textcolor{red}{$\ominus$} No pure localization mode \newline
    \textcolor{red}{$\ominus$} No multi-session or multi-robot support
  & \textcolor{green}{$\oplus$} Fast map updates via multi-threading \newline
    \textcolor{red}{$\ominus$} Slow ORB feature extraction \newline
    \textcolor{red}{$\ominus$} Inference time degrades with map size
  & \textcolor{green}{$\oplus$} Competitive accuracy in industrial settings \newline
    \textcolor{red}{$\ominus$} Heavy development effort required \newline
    \textcolor{red}{$\ominus$} Limited production support
  & 4\textsuperscript{th} \\
\bottomrule
\end{tabular}}
\end{table}

\section{Conclusion}
\label{sec:conclusion}

This study presents a comprehensive evaluation of visual odometry and SLAM systems for autonomous mobile robots in logistics environments. Our method leverages NVIDIA's GPU-accelerated cuVSLAM stereo pipeline, where the power of NVIDIA GPUs enables real-time, robust visual odometry that significantly outperforms state-of-the-art alternatives across diverse motion profiles. The results underscore how NVIDIA's hardware and software ecosystem unlocks a new level of localization accuracy and robustness, paving the way for more reliable and efficient autonomous navigation in logistics applications. Future work will focus on extending the cuVSLAM pipeline with RGB-D sensor support, further broadening the applicability of NVIDIA's visual SLAM stack to a wider range of robot platforms and deployment scenarios. With the cuVSLAM source code now open‑sourced on GitHub (https://github.com/nvidia-isaac/cuVSLAM), we are looking forward to contributing to its continued evolution to accelerate the maturity of GPU‑accelerated VSLAM for industrial use.

\bibliographystyle{plainnat}  
\bibliography{main.bib}

\end{document}